\documentclass[10pt,twocolumn,letterpaper]{article}
\pdfoutput=1
\usepackage{cvpr}
\usepackage{times}
\usepackage{epsfig}
\usepackage{graphicx}
\usepackage{amsmath}
\usepackage{amssymb}
\usepackage{multirow}

\newcommand{\denselist}{\itemsep 0pt\parsep=0pt\partopsep0pt\vspace{-\topsep}}

\newcommand{\bitem}{\begin{itemize}\denselist}
	\newcommand{\eitem}{\end{itemize}}
\newcommand{\benum}{\begin{enumerate}\denselist}
	\newcommand{\eenum}{\end{enumerate}}
\newcommand{\bdescr}{\begin{description}\denselist}
	\newcommand{\edescr}{\end{description}}

\usepackage[pagebackref=true,breaklinks=true,letterpaper=true,colorlinks,bookmarks=false]{hyperref}

\cvprfinalcopy 

\def\cvprPaperID{19} 

\ifcvprfinal\fi
\begin{document}

\title{Frustum PointNets for 3D Object Detection from RGB-D Data}

\author{Charles R. Qi$^1$\thanks{Majority of the work done as an intern at Nuro, Inc.}\qquad Wei Liu$^2$\qquad Chenxia Wu$^2$\qquad Hao Su$^3$\qquad Leonidas J. Guibas$^1$\\
$^1$Stanford University\qquad $^2$Nuro, Inc.\qquad $^3$UC San Diego}

\maketitle

\begin{abstract}


In this work, we study 3D object detection from RGB-D data in both indoor and outdoor scenes. While previous methods focus on images or 3D voxels, often obscuring natural 3D patterns and invariances of 3D data, we directly operate on raw point clouds by popping up RGB-D scans. However, a key challenge of this approach is how to efficiently localize objects in point clouds of large-scale scenes (region proposal). Instead of solely relying on 3D proposals, our method leverages both mature 2D object detectors and advanced 3D deep learning for object localization, achieving efficiency as well as high recall for even small objects. Benefited from learning directly in raw point clouds, our method is also able to precisely estimate 3D bounding boxes even under strong occlusion or with very sparse points. Evaluated on KITTI and SUN RGB-D 3D detection benchmarks, our method outperforms the state of the art by remarkable margins while having real-time capability.

\end{abstract}

\section{Introduction}
\label{sec:intro}


Recently, great progress has been made on 2D image understanding tasks, such as object detection~\cite{girshick2014rich} and instance segmentation~\cite{he2017mask}. However, beyond getting 2D bounding boxes or pixel masks, \emph{3D understanding} is eagerly in demand in many applications such as autonomous driving and augmented reality (AR). With the popularity of 3D sensors deployed on mobile devices and autonomous vehicles, more and more 3D data is captured and processed. In this work, we study one of the most important 3D perception tasks -- 3D object detection, which classifies the object category and estimates \emph{oriented 3D bounding boxes} of physical objects from 3D sensor data. 


While 3D sensor data is often in the form of point clouds, how to represent point cloud and what deep net architectures to use for 3D object detection remains an open problem. Most existing works convert 3D point clouds to images by projection~\cite{su15mvcnn,qi2016volumetric} or to volumetric grids by quantization~\cite{wu20153d,maturana2015voxnet,qi2016volumetric} and then apply convolutional networks. This data representation transformation, however, may obscure natural 3D patterns and invariances of the data. 
Recently, a number of papers have proposed to process point clouds directly without converting them to other formats. For example, \cite{qi2017pointnet, qi2017pointnetplusplus} proposed new types of deep net architectures, called \emph{PointNets}, which have shown superior performance and efficiency in several 3D understanding tasks such as object classification and semantic segmentation.

\begin{figure}[t!]
    \centering
    \includegraphics[width=\linewidth]{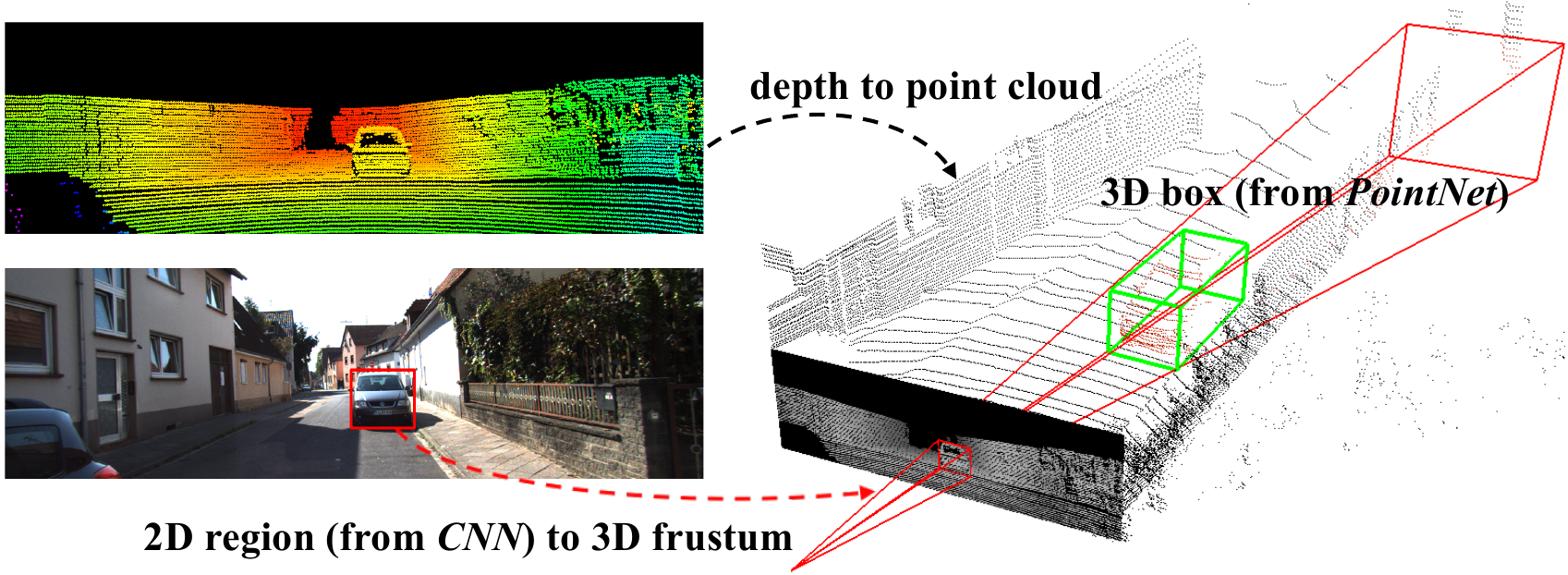}
    \caption{\textbf{3D object detection pipeline.} Given RGB-D data, we first generate 2D object region proposals in the RGB image using a CNN. Each 2D region is then extruded to a \emph{3D viewing frustum} in which we get a point cloud from depth data. Finally, our frustum PointNet predicts a (oriented and amodal) 3D bounding box for the object from the points in frustum.}
    \label{fig:teaser}
\end{figure}

While PointNets are capable of classifying a whole point cloud or predicting a semantic class for each point in a point cloud, it is unclear how this architecture can be used for instance-level 3D object detection. Towards this goal, we have to address one key challenge: how to efficiently propose possible locations of 3D objects in a 3D space. Imitating the practice in image detection, it is straightforward to enumerate candidate 3D boxes by sliding windows~\cite{engelcke2017vote3deep} or by 3D region proposal networks such as \cite{song2015sun}. However, the computational complexity of 3D search typically grows cubically with respect to resolution and becomes too expensive for large scenes or real-time applications such as autonomous driving. 

Instead, in this work, we reduce the search space following the dimension reduction principle: we take the advantage of mature 2D object detectors (Fig.~\ref{fig:teaser}). First, we extract the 3D bounding frustum of an object by extruding 2D bounding boxes from image detectors. Then, within the 3D space trimmed by each of the 3D frustums, we consecutively perform 3D object instance segmentation and \emph{amodal} 3D bounding box regression using two variants of PointNet. The segmentation network predicts the 3D mask of the object of interest (i.e. instance segmentation); and the regression network estimates the amodal 3D bounding box (covering the entire object even if only part of it is visible).

In contrast to previous work that treats RGB-D data as 2D maps for CNNs, our method is more \emph{3D-centric} as we lift depth maps to 3D point clouds and process them using 3D tools. This 3D-centric view enables new capabilities for exploring 3D data in a more effective manner. First, in our pipeline, a few transformations are applied successively on 3D coordinates, which align point clouds into a sequence of more constrained and canonical frames.
These alignments factor out pose variations in data, and thus make 3D geometry pattern more evident, leading to an easier job of 3D learners.
Second, learning in 3D space can better exploits the geometric and topological structure of 3D space.
In principle, all objects live in 3D space; therefore, we believe that many geometric structures, such as repetition, planarity, and symmetry, are more naturally parameterized and captured by learners that directly operate in 3D space. 
The usefulness of this 3D-centric network design philosophy has been supported by much recent experimental evidence.

Our method achieve leading positions on KITTI 3D object detection~\cite{kitti-3d-detection} and bird's eye view detection~\cite{kitti-3d-localization} benchmarks.
Compared with the previous state of the art~\cite{cvpr17chen}, our method is \emph{8.04\%} better on 3D car AP with high efficiency (running at 5 fps). Our method also fits well to indoor RGB-D data where we have achieved \emph{8.9\%} and \emph{6.4\%} better 3D mAP than ~\cite{lahoud20172d} and ~\cite{ren2016three} on SUN-RGBD while running one to three orders of magnitude faster.


The key contributions of our work are as follows:
\bitem
    \item We propose a novel framework for RGB-D data based 3D object detection called Frustum PointNets.
    \item We show how we can train 3D object detectors under our framework and achieve state-of-the-art performance on standard 3D object detection benchmarks.
    \item We provide extensive quantitative evaluations to validate our design choices as well as rich qualitative results for understanding the strengths and limitations of our method.
\eitem

\begin{figure*}[t!]
    \centering
    \includegraphics[width=0.92\linewidth]{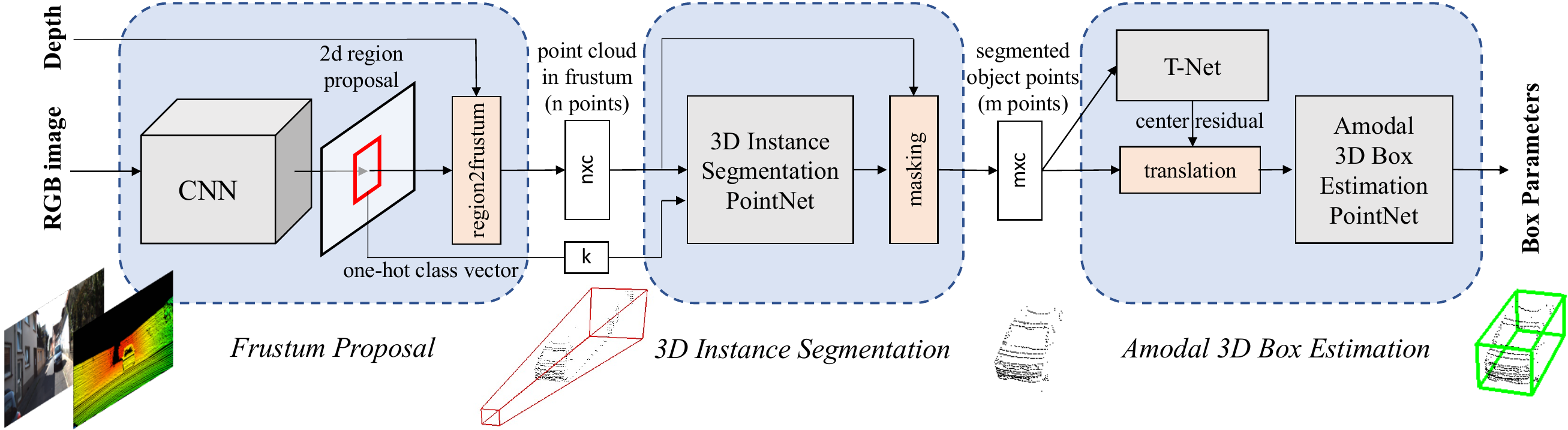}
    \caption{\textbf{Frustum PointNets for 3D object detection.} We first leverage a 2D CNN object detector to propose 2D regions and classify their content. 2D regions are then lifted to 3D and thus become frustum proposals. Given a point cloud in a frustum ($n\times c$ with $n$ points and $c$ channels of XYZ, intensity etc. for each point), the object instance is segmented by binary classification of each point. Based on the segmented object point cloud ($m \times c$), a light-weight regression PointNet (T-Net) tries to align points by translation such that their centroid is close to amodal box center. At last the box estimation net estimates the amodal 3D bounding box for the object. More illustrations on coordinate systems involved and network input, output are in Fig.~\ref{fig:coordinate} and Fig.~\ref{fig:network}.}
    \label{fig:pipeline}
\end{figure*}

\section{Related Work}
\paragraph{3D Object Detection from RGB-D Data} Researchers have approached the 3D detection problem by taking various ways to represent RGB-D data.

\emph{Front view image based methods:} ~\cite{chen2016monocular, mousavian20163d, xiang2015data} take monocular RGB images and shape priors or occlusion patterns to infer 3D bounding boxes. ~\cite{li2016vehicle, deng2017amodal} represent depth data as 2D maps and apply CNNs to localize objects in 2D image. In comparison we represent depth as a point cloud and use advanced 3D deep networks (PointNets) that can exploit 3D geometry more effectively.

\emph{Bird's eye view based methods:} MV3D~\cite{cvpr17chen} projects LiDAR point cloud to bird's eye view and trains a region proposal network (RPN~\cite{ren2015faster}) for 3D bounding box proposal. However, the method lags behind in detecting small objects, such as pedestrians and cyclists and cannot easily adapt to scenes with multiple objects in vertical direction.

\emph{3D based methods:} ~\cite{wang2015voting, song2014sliding} train 3D object classifiers by SVMs on hand-designed geometry features extracted from point cloud and then localize objects using sliding-window search. \cite{engelcke2017vote3deep} extends ~\cite{wang2015voting} by replacing SVM with 3D CNN on voxelized 3D grids. \cite{ren2016three} designs new geometric features for 3D object detection in a point cloud. \cite{song2016deep, li20163d} convert a point cloud of the entire scene into a volumetric grid and use 3D volumetric CNN for object proposal and classification. Computation cost for those method is usually quite high due to the expensive cost of 3D convolutions and large 3D search space.
Recently, \cite{lahoud20172d} proposes a 2D-driven 3D object detection method that is similar to ours in spirit. However, they use hand-crafted features (based on histogram of point coordinates) with simple fully connected networks to regress 3D box location and pose, which is sub-optimal in both speed and performance. In contrast, we propose a more flexible and effective solution with deep 3D feature learning (PointNets).

\paragraph{Deep Learning on Point Clouds}
Most existing works convert point clouds to images or volumetric forms before feature learning. \cite{wu20153d, maturana2015voxnet, qi2016volumetric} voxelize point clouds into volumetric grids and generalize image CNNs to 3D CNNs. ~\cite{li2016fpnn, riegler2016octnet, wang2017cnn, engelcke2017vote3deep} design more efficient 3D CNN or neural network architectures that exploit sparsity in point cloud.
However, these CNN based methods still require quantitization of point clouds with certain voxel resolution.
Recently, a few works~\cite{qi2017pointnet,qi2017pointnetplusplus} propose a novel type of network architectures (PointNets) that directly consumes raw point clouds without converting them to other formats. While PointNets have been applied to single object classification and semantic segmentation, our work explores how to extend the architecture for the purpose of 3D object detection.

\section{Problem Definition}
\label{sec:problem_definition}
Given RGB-D data as input, our goal is to classify and localize objects in 3D space. The depth data, obtained from LiDAR or indoor depth sensors, is represented as a point cloud in RGB camera coordinates. The projection matrix is also known so that we can get a 3D frustum from a 2D image region. Each object is represented by a class (one among $k$ predefined classes) and an \emph{amodal} 3D bounding box. The \emph{amodal} box bounds the complete object even if part of the object is occluded or truncated. The 3D box is parameterized by its size $h,w,l$, center $c_x, c_y, c_z$, and orientation $\theta, \phi, \psi$ relative to a predefined canonical pose for each category. In our implementation, we only consider the heading angle $\theta$ around the up-axis for orientation.

\section{3D Detection with Frustum PointNets}
As shown in Fig.~\ref{fig:pipeline}, our system for 3D object detection consists of three modules: frustum proposal, 3D instance segmentation, and 3D amodal bounding box estimation. We will introduce each module in the following subsections. We will focus on the pipeline and functionality of each module, and refer readers to supplementary for specific architectures of the deep networks involved.

\subsection{Frustum Proposal}
\label{sec:frustum_proposal}

The resolution of data produced by most 3D sensors, especially real-time depth sensors, is still lower than RGB images from commodity cameras. Therefore, we leverage mature 2D object detector to propose 2D object regions in RGB images as well as to classify objects.

\begin{figure}[t!]
    \centering
    \includegraphics[width=0.95\linewidth]{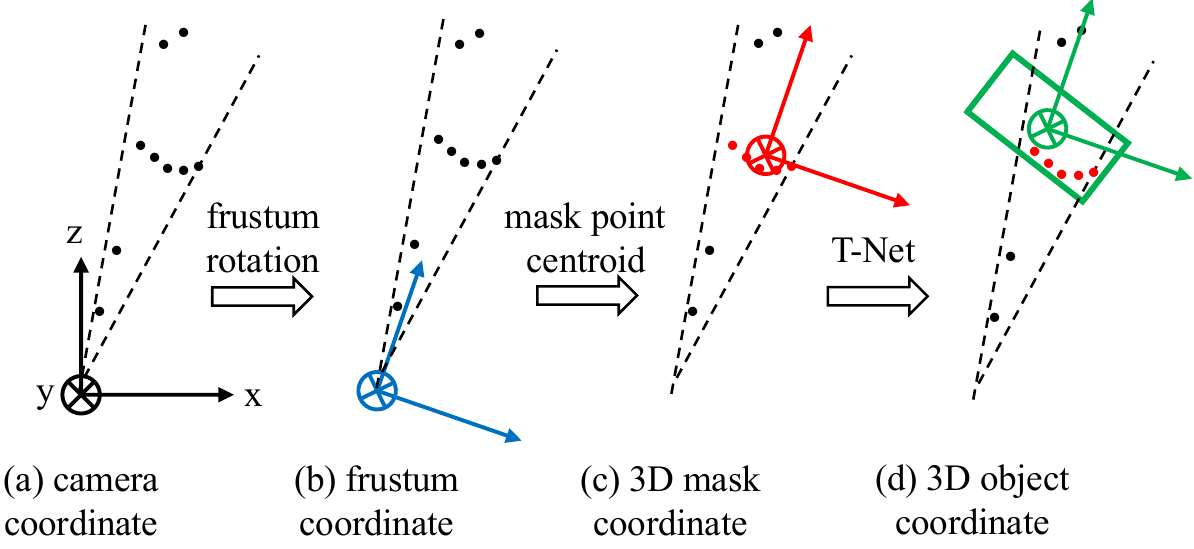}
    \caption{\textbf{Coordinate systems for point cloud.} Artificial points (black dots) are shown to illustrate (a) default camera coordinate; (b) frustum coordinate after rotating frustums to center view (Sec.~\ref{sec:frustum_proposal}); (c) mask coordinate with object points' centroid at origin (Sec.~\ref{sec:instance_seg}); (d) object coordinate predicted by T-Net (Sec.~\ref{sec:box_estimation}).
    }
    \label{fig:coordinate}
\end{figure}

With a known camera projection matrix, a 2D bounding box can be lifted to a frustum (with near and far planes specified by depth sensor range) that defines a 3D search space for the object. We then collect all points within the frustum to form a \emph{frustum point cloud}. As shown in Fig~\ref{fig:coordinate} (a), frustums may orient towards many different directions, which result in large variation in the placement of point clouds.  We therefore normalize the frustums by rotating them toward a center view such that the center axis of the frustum is orthogonal to the image plane. This normalization helps improve the rotation-invariance of the algorithm. 
We call this entire procedure for extracting frustum point clouds from RGB-D data \emph{frustum proposal generation}.

\begin{figure}[b!]
    \centering
    \includegraphics[width=0.8\linewidth]{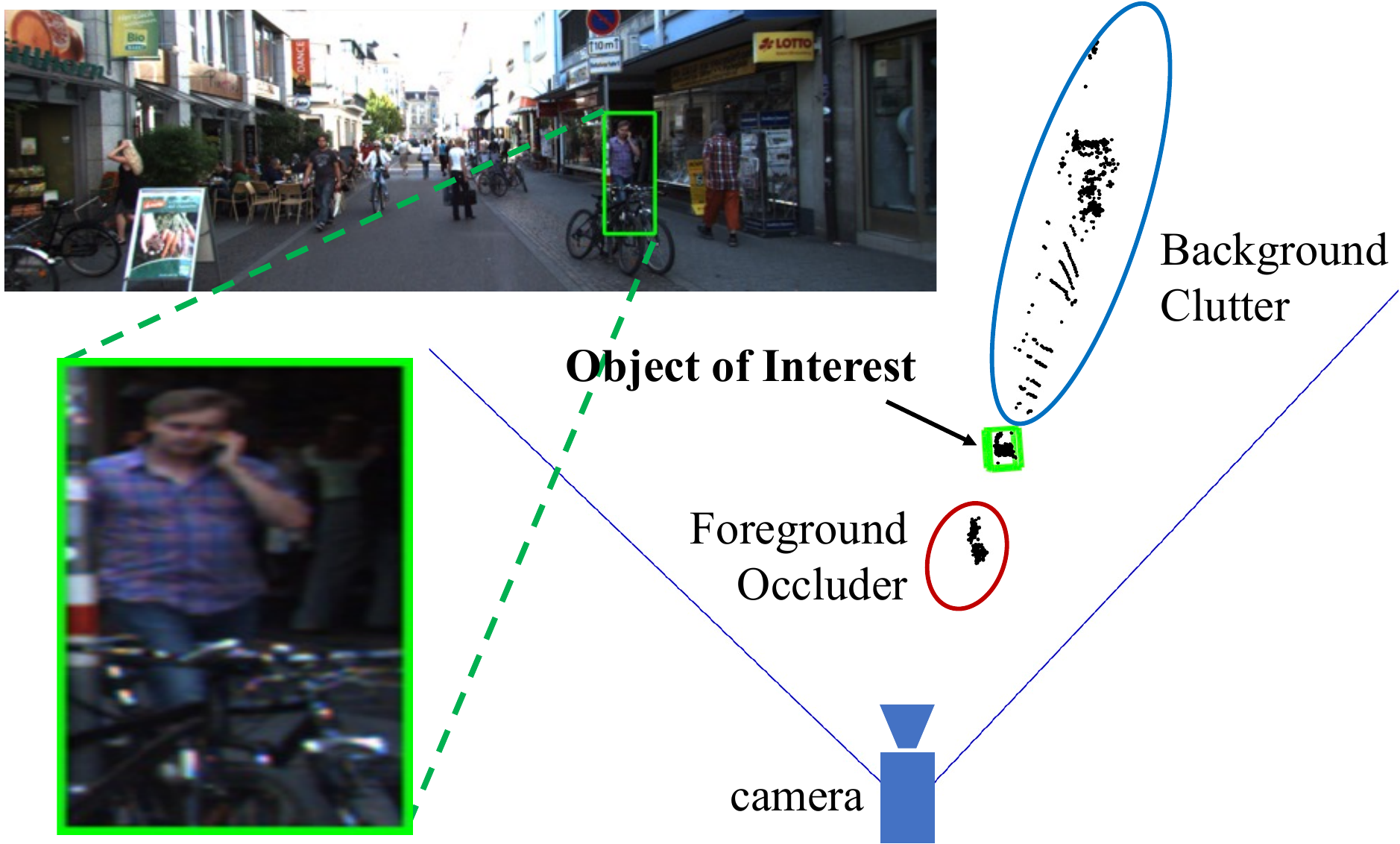}
    \caption{\textbf{Challenges for 3D detection in frustum point cloud.} \emph{Left:} RGB image with an image region proposal for a person. \emph{Right:} bird's eye view of the LiDAR points in the extruded frustum from 2D box, where we see a wide spread of points with both foreground occluder (bikes) and background clutter (building).}
    \label{fig:frustum}
\end{figure}


While our 3D detection framework is agnostic to the exact method for 2D region proposal, we adopt a FPN~\cite{lin2016feature} based model. We pre-train the model weights on ImageNet classification and COCO object detection datasets and further fine-tune it on a KITTI 2D object detection dataset to classify and predict \emph{amodal} 2D boxes. More details of the 2D detector training are provided in the supplementary.

\subsection{3D Instance Segmentation}
\label{sec:instance_seg}

Given a 2D image region (and its corresponding 3D frustum), several methods might be used to obtain 3D location of the object:
One straightforward solution is to directly regress 3D object locations (e.g., by 3D bounding box) from a depth map using 2D CNNs. However, this problem is not easy as occluding objects and background clutter is common in natural scenes (as in Fig.~\ref{fig:frustum}), which may severely distract the 3D localization task. 
Because objects are naturally separated in physical space, segmentation in 3D point cloud is much more natural and easier than that in images where pixels from distant objects can be near-by to each other. Having observed this fact, we propose to segment instances in 3D point cloud instead of in 2D image or depth map. Similar to Mask-RCNN~\cite{he2017mask}, which achieves instance segmentation by binary classification of pixels in image regions, we realize 3D instance segmentation using a PointNet-based network on point clouds in frustums.

Based on 3D instance segmentation, we are able to achieve \emph{residual} based 3D localization. That is, rather than regressing the absolute 3D location of the object whose offset from the sensor may vary in large ranges (e.g. from 5m to beyond 50m in KITTI data), we predict the 3D bounding box center in a local coordinate system -- 3D mask coordinates as shown in Fig.~\ref{fig:coordinate} (c).


\paragraph{3D Instance Segmentation PointNet.} The network takes a point cloud in frustum and predicts a probability score for each point that indicates how likely the point belongs to the object of interest. Note that each frustum contains exactly one object of interest. Here those ``other'' points could be points of non-relevant areas (such as ground, vegetation) or other instances that occlude or are behind the object of interest. Similar to the case in 2D instance segmentation, depending on the position of the frustum, object points in one frustum may become cluttered or occlude points in another. Therefore, our segmentation PointNet is learning the occlusion and clutter patterns as well as recognizing the geometry for the object of a certain category.


In a multi-class detection case, we also leverage the semantics from a 2D detector for better instance segmentation. For example, if we know the object of interest is a pedestrian, then the segmentation network can use this prior to find geometries that look like a person. Specifically, in our architecture we encode the semantic category as a one-hot class vector ($k$ dimensional for the pre-defined $k$ categories) and concatenate the one-hot vector to the intermediate point cloud features. More details of the specific architectures are described in the supplementary.

After 3D instance segmentation, points that are classified as the object of interest are extracted (``masking'' in Fig.~\ref{fig:pipeline}). Having obtained these segmented object points, we further normalize its coordinates to boost the translational invariance of the algorithm, following the same rationale as in the frustum proposal step.  In our implementation, we transform the point cloud into a local coordinate by subtracting XYZ values by its centroid. This is illustrated in Fig.~\ref{fig:coordinate} (c). Note that we intentionally do not scale the point cloud, because the bounding sphere size of a partial point cloud can be greatly affected by viewpoints and the real size of the point cloud helps the box size estimation.

In our experiments, we find that coordinate transformations such as the one above and the previous frustum rotation are critical for 3D detection result as shown in Tab.~\ref{tab:pc_normalization}.  


\begin{figure}[t!]
    \centering
    \includegraphics[width=\linewidth]{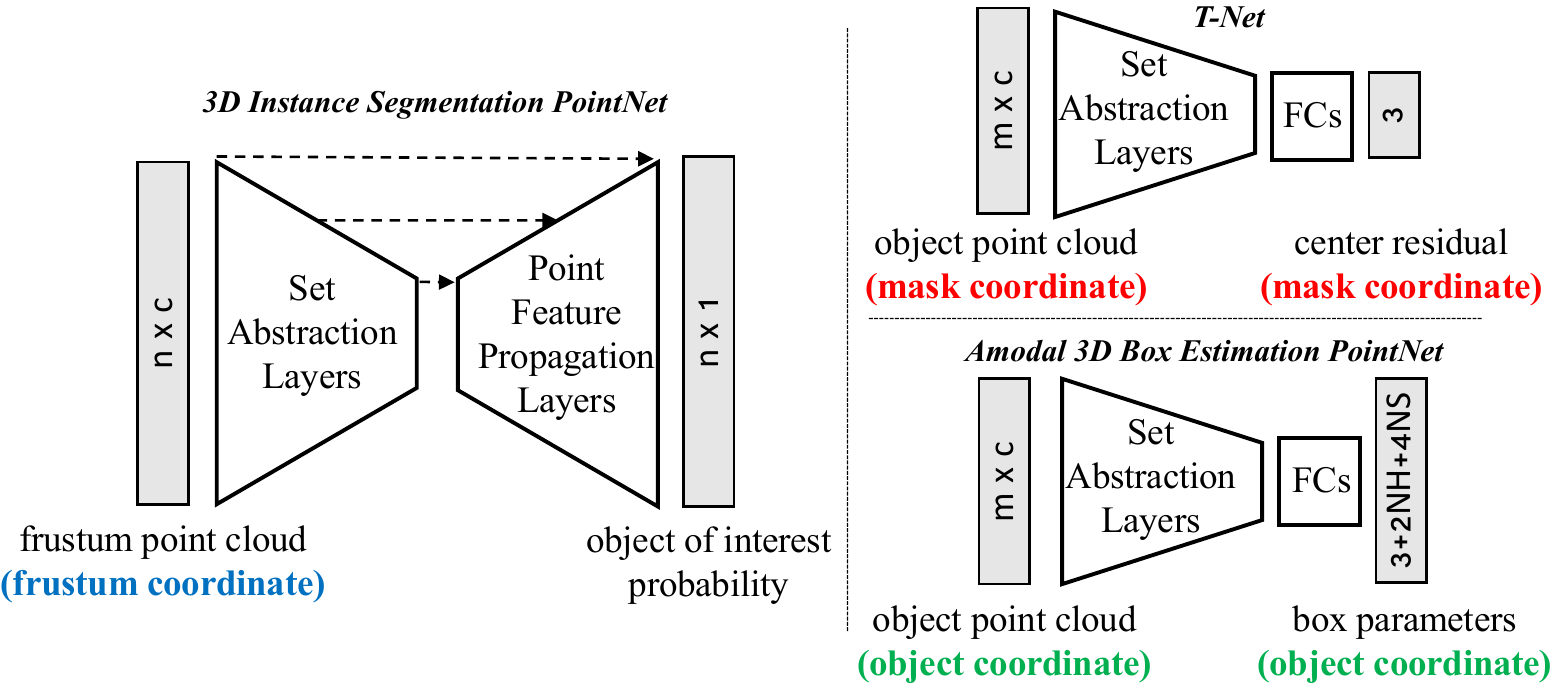}
    \caption{\textbf{Basic architectures and IO for PointNets.} Architecture is illustrated for PointNet++~\cite{qi2017pointnetplusplus} (v2) models with set abstraction layers and feature propagation layers (for segmentation). Coordinate systems involved are visualized in Fig.~\ref{fig:coordinate}.}
    \label{fig:network}
\end{figure}

\subsection{Amodal 3D Box Estimation}
\label{sec:box_estimation}

Given the segmented object points (in 3D mask coordinate), this module estimates the object's amodal oriented 3D bounding box by using a box regression PointNet together with a preprocessing transformer network.

\paragraph{Learning-based 3D Alignment by T-Net} Even though we have aligned segmented object points according to their centroid position, we find that the origin of the mask coordinate frame (Fig.~\ref{fig:coordinate} (c)) may still be quite far from the \emph{amodal} box center. We therefore propose to use a light-weight regression PointNet (T-Net) to estimate the true center of the complete object and then transform the coordinate such that the predicted center becomes the origin (Fig.~\ref{fig:coordinate} (d)).

The architecture and training of our T-Net is similar to the T-Net in \cite{qi2017pointnet}, which can be thought of as a special type of spatial transformer network (STN)~\cite{jaderberg2015spatial}. However, different from the original STN that has no direct supervision on transformation, we explicitly supervise our translation network to predict center residuals from the mask coordinate origin to real object center. 

\paragraph{Amodal 3D Box Estimation PointNet}
The box estimation network predicts amodal bounding boxes (for entire object even if part of it is unseen) for objects given an object point cloud in 3D object coordinate (Fig.~\ref{fig:coordinate} (d)). The network architecture is similar to that for object classification~\cite{qi2017pointnet, qi2017pointnetplusplus}, however the output is no longer object class scores but parameters for a 3D bounding box.

As stated in Sec.~\ref{sec:problem_definition}, we parameterize a 3D bounding box by its center ($c_x$, $c_y$, $c_z$), size ($h$, $w$, $l$) and heading angle $\theta$ (along up-axis). We take a ``residual'' approach for box center estimation. The center residual predicted by the box estimation network is combined with the previous center residual from the T-Net and the masked points' centroid to recover an absolute center (Eq.~\ref{eq:center}). For box size and heading angle, we follow previous works~\cite{ren2015faster, mousavian20163d} and use a hybrid of classification and regression formulations. Specifically we pre-define $NS$ size templates and $NH$ equally split angle bins. Our model will both classify size/heading ($NS$ scores for size, $NH$ scores for heading) to those pre-defined categories as well as predict residual numbers for each category ($3\times NS$ residual dimensions for height, width, length, $NH$ residual angles for heading). In the end the net outputs $3+4\times NS+2\times NH$ numbers in total.

\begin{equation}
    \label{eq:center}
    C_{pred} = C_{mask} + \Delta C_{t-net} + \Delta C_{box-net}
\end{equation}



\begin{table*}[t!]
\small
\centering
\begin{tabular}{l||ccc||ccc||ccc}
\hline
\multirow{2}{*}{Method} & \multicolumn{3}{c||}{Cars} & \multicolumn{3}{c||}{Pedestrians} & \multicolumn{3}{c}{Cyclists} \\ \cline{2-10} 
                        & Easy  & Moderate  & Hard  & Easy     & Moderate    & Hard    & Easy    & Moderate   & Hard   \\ \hline
DoBEM~\cite{yuvehicle} & 7.42 & 6.95 & 13.45 & - & - & - & - & - & - \\
MV3D~\cite{cvpr17chen} & 71.09 & 62.35 & 55.12 & - & - & - & - & - & - \\ \hline
Ours (v1) & 80.62 & 64.70 & 56.07 & 50.88 & 41.55 & 38.04 & 69.36 & 53.50 & 52.88 \\
Ours (v2) & \textbf{81.20} & \textbf{70.39} & \textbf{62.19} & \textbf{51.21} & \textbf{44.89} & \textbf{40.23} & \textbf{71.96} & \textbf{56.77} & \textbf{50.39} \\ \hline
\end{tabular}
\caption{\textbf{3D object detection} \emph{3D} AP on KITTI \emph{test} set. DoBEM~\cite{yuvehicle} and MV3D~\cite{cvpr17chen} (previous state of the art) are based on 2D CNNs with bird's eye view LiDAR image.
Our method, without sensor fusion or multi-view aggregation, outperforms those methods by large margins on all categories and data subsets. 3D bounding box IoU threshold is 70\% for cars and 50\% for pedestrians and cyclists.}
\label{tab:kitti_test_3d_detection}
\end{table*}

\begin{table*}[t!]
\small
\centering
\begin{tabular}{l||ccc||ccc||ccc}
\hline
\multirow{2}{*}{Method} & \multicolumn{3}{c||}{Cars} & \multicolumn{3}{c||}{Pedestrians} & \multicolumn{3}{c}{Cyclists} \\ \cline{2-10} 
                        & Easy  & Moderate  & Hard  & Easy     & Moderate    & Hard    & Easy    & Moderate   & Hard   \\ \hline
DoBEM~\cite{yuvehicle} & 36.49 & 36.95 & 38.10 & - & - & - & - & - & - \\
3D FCN~\cite{li20163d} & 69.94 & 62.54 & 55.94 & - & - & - & - & - & - \\
MV3D~\cite{cvpr17chen} & 86.02 & 76.90 & 68.49 & - & - & - & - & - & - \\ \hline
Ours (v1) & 87.28 & 77.09 & 67.90 & 55.26 & 47.56 & 42.57 & 73.42 & 59.87 & 52.88 \\
Ours (v2) & \textbf{88.70} & \textbf{84.00} & \textbf{75.33} & \textbf{58.09} & \textbf{50.22} & \textbf{47.20} & \textbf{75.38} & \textbf{61.96} & \textbf{54.68} \\ \hline
\end{tabular}
\caption{\textbf{3D object localization} AP (bird's eye view) on KITTI \emph{test} set.
3D FCN~\cite{li20163d} uses 3D CNNs on voxelized point cloud and is far from real-time. MV3D~\cite{cvpr17chen} is the previous state of the art.
Our method significantly outperforms those methods on all categories and data subsets. Bird's eye view 2D bounding box IoU threshold is 70\% for cars and 50\% for pedestrians and cyclists.}
\label{tab:kitti_test_3d_localization}

\end{table*}

\subsection{Training with Multi-task Losses}
\label{sec:losses}

We simultaneously optimize the three nets involved (3D instance segmentation PointNet, T-Net and amodal box estimation PointNet) with multi-task losses (as in Eq.~\ref{eq:losses}).
$L_{c1-reg}$ is for T-Net and $L_{c2-reg}$ is for center regression of box estimation net. $L_{h-cls}$ and $L_{h-reg}$ are losses for heading angle prediction while $L_{s-cls}$ and $L_{s-reg}$ are for box size.
Softmax is used for all classification tasks and smooth-$l_1$ (huber) loss is used for all regression cases.

\vspace{-1mm}
\begin{equation}
\small
\label{eq:losses}
\begin{split}
    L_{multi-task} = & L_{seg} + \lambda (L_{c1-reg} + L_{c2-reg} + L_{h-cls} + \\
    & L_{h-reg} + L_{s-cls} + L_{s-reg} + \gamma L_{corner})
\end{split}
\end{equation}

\paragraph{Corner Loss for Joint Optimization of Box Parameters}
While our 3D bounding box parameterization is compact and complete, learning is not optimized for final 3D box accuracy -- center, size and heading have \emph{separate} loss terms. Imagine cases where center and size are accurately predicted but heading angle is off -- the 3D IoU with ground truth box will then be dominated by the angle error. Ideally all three terms (center,size,heading) should be \emph{jointly} optimized for best 3D box estimation (under IoU metric).
To resolve this problem we propose a novel regularization loss, the \emph{corner loss}:

\vspace{-1mm}

\begin{equation} 
\small
\label{eq:corner_loss}
    L_{corner} = \sum_{i=1}^{NS} \sum_{j=1}^{NH} \delta_{ij} \text{min} \{ \sum_{k=1}^{8} \| P_{k}^{ij} - P_{k}^{*}\|, \sum_{i=1}^{8} \| P_{k}^{ij} - P_{k}^{**} \| \}
\end{equation}

In essence, the corner loss is the sum of the distances between the eight corners of a predicted box and a ground truth box. Since corner positions are jointly determined by center, size and heading, the corner loss is able to regularize the multi-task training for those parameters.

To compute the corner loss, we firstly construct $NS \times NH$ ``anchor'' boxes from all size templates and heading angle bins. The anchor boxes are then translated to the estimated box center. We denote the anchor box corners as $P_k^{ij}$, where $i$, $j$, $k$ are indices for the size class, heading class, and (predefined) corner order, respectively.
To avoid large penalty from flipped heading estimation, we further compute distances to corners ($P_k^{**})$ from the flipped ground truth box and use the minimum of the original and flipped cases. $\delta_{ij}$, which is one for the ground truth size/heading class and zero else wise, is a two-dimensional mask used to select the distance term we care about.

\section{Experiments}
Experiments are divided into three parts\footnote{Details on network architectures, training parameters as well as more experiments are included in the supplementary material.}. First we compare with state-of-the-art methods for 3D object detection on KITTI~\cite{geiger2013vision} and SUN-RGBD~\cite{song2015sun} (Sec~\ref{sec:exp_benchmark}). Second, we provide in-depth analysis to validate our design choices (Sec~\ref{sec:exp_analysis}). Last, we show qualitative results and discuss the strengths and limitations of our methods (Sec~\ref{sec:exp_viz}).


\subsection{Comparing with state-of-the-art Methods} \label{sec:exp_benchmark}
We evaluate our 3D object detector on KITTI~\cite{Geiger2012CVPR} and SUN-RGBD~\cite{song2015sun} benchmarks for 3D object detection. On both tasks we have achieved significantly better results compared with state-of-the-art methods.


\paragraph{KITTI} Tab.~\ref{tab:kitti_test_3d_detection} shows the performance of our 3D detector on the KITTI \emph{test} set. We outperform previous state-of-the-art methods by a large margin. While MV3D~\cite{cvpr17chen} uses multi-view feature aggregation and sophisticated multi-sensor fusion strategy, our method based on the PointNet~\cite{qi2017pointnet} (v1) and PointNet++~\cite{qi2017pointnetplusplus} (v2) backbone is much cleaner in design. While out of the scope for this work, we expect that sensor fusion (esp. aggregation of image feature for 3D detection) could further improve our results.

We also show our method's performance on 3D object localization (bird's eye view) in Tab.~\ref{tab:kitti_test_3d_localization}. In the 3D localization task bounding boxes are projected to bird's eye view plane and IoU is evaluated on oriented 2D boxes. Again, our method significantly outperforms previous works which include DoBEM~\cite{yuvehicle} and MV3D~\cite{cvpr17chen} that use CNNs on projected LiDAR images, as well as 3D FCN~\cite{li20163d} that uses 3D CNNs on voxelized point cloud. 

\begin{figure*}[t!]
    \centering
    \includegraphics[width=\linewidth]{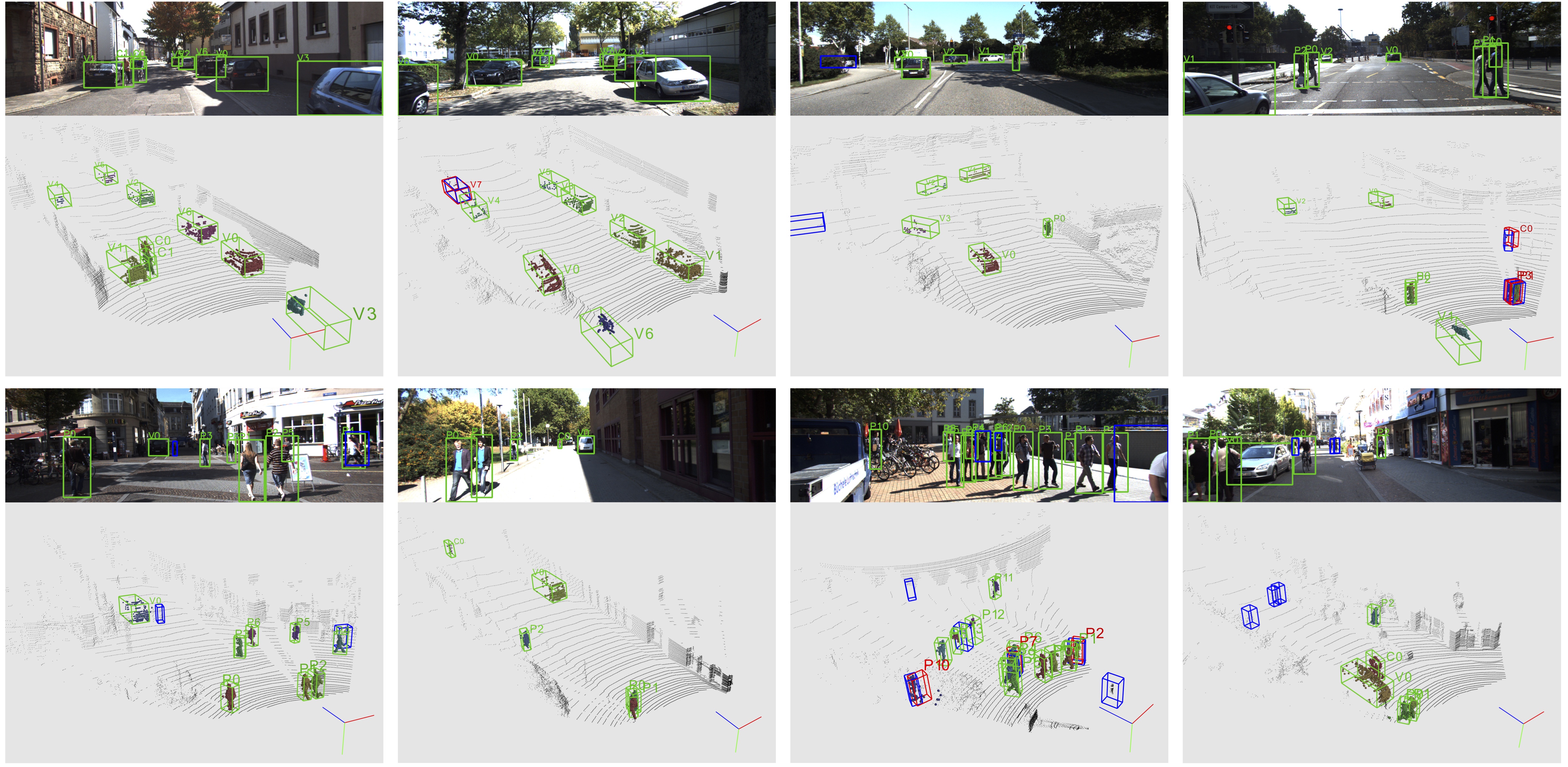}
    \caption{\textbf{Visualizations of Frustum PointNet results on KITTI val set}  (best viewed in color with zoom in). These results are based on PointNet++ models~\cite{qi2017pointnetplusplus}, running at 5 fps and achieving test set \emph{3D} AP of 70.39, 44.89 and 56.77 for car, pedestrian and cyclist, respectively. 3D instance masks on point cloud are shown in color. True positive detection boxes are in green, while false positive boxes are in red and groundtruth boxes in blue are shown for false positive and false negative cases. Digit and letter beside each box denote instance id and semantic class, with ``v'' for cars, ``p'' for pedestrian and ``c'' for cyclist. See Sec.~\ref{sec:exp_viz} for more discussion on the results.}
    \label{fig:results}
\end{figure*}

The output of our network is visualized in Fig.~\ref{fig:results} where we observe accurate 3D instance segmentation and box prediction even under very challenging cases. We defer more discussions on success and failure case patterns to Sec.~\ref{sec:exp_viz}. We also report performance on KITTI val set (the same split as in~\cite{cvpr17chen}) in Tab.~\ref{tab:kitti_val_3d_detection} and Tab.~\ref{tab:kitti_val_3d_localization} (for cars) to support comparison with more published works, and in Tab.~\ref{tab:val_ped_cyc} (for pedestrians and cyclists) for reference.


\begin{table}[t!]
\small
\centering
\label{tab:kitti_valid}
\begin{tabular}{l||ccc}
\hline
Method & Easy    & Moderate    & Hard   \\ \hline
Mono3D~\cite{chen2016monocular} & 2.53 & 2.31 & 2.31 \\ 
3DOP~\cite{chen20153d}  & 6.55 & 5.07 & 4.10 \\ \hline
VeloFCN~\cite{li20163d} & 15.20 & 13.66 & 15.98 \\ 
MV3D (LiDAR)~\cite{cvpr17chen} & 71.19 & 56.60 & 55.30 \\ 
MV3D~\cite{cvpr17chen} & 71.29 & 62.68 & 56.56 \\ \hline
Ours (v1) & 83.26 & 69.28 & 62.56 \\
Ours (v2) & \textbf{83.76} & \textbf{70.92} & \textbf{63.65}\\ \hline
\end{tabular}
\caption{\textbf{3D object detection} AP on KITTI \emph{val} set (cars only).}
\label{tab:kitti_val_3d_detection}
\end{table}

\begin{table}[t!]
\small
\centering
\label{tab:kitti_valid}
\begin{tabular}{l||ccc}
\hline
Method & Easy    & Moderate   & Hard   \\ \hline
Mono3D~\cite{chen2016monocular} & 5.22 & 5.19 & 4.13 \\ 
3DOP~\cite{chen20153d}  & 12.63 & 9.49 & 7.59 \\ \hline
VeloFCN~\cite{li20163d} &  40.14 & 32.08 & 30.47 \\ 
MV3D (LiDAR)~\cite{cvpr17chen} & 86.18 & 77.32 & 76.33 \\ 
MV3D~\cite{cvpr17chen} & 86.55 & 78.10 & \textbf{76.67} \\ \hline
Ours (v1) & 87.82 & 82.44 & 74.77 \\
Ours (v2) & \textbf{88.16} & \textbf{84.02} & 76.44 \\ \hline
\end{tabular}
\caption{\textbf{3D object localization} AP on KITTI \emph{val} set (cars only).}
\label{tab:kitti_val_3d_localization}
\end{table}

\begin{table}[h!]
\small
\centering
\begin{tabular}{|c|c|c|c|}
    \hline
    Benchmark & Easy & Moderate & Hard \\ \hline
    Pedestrian (3D Detection) & 70.00 & 61.32 & 53.59 \\ \hline
    Pedestrian (Bird's Eye View) & 72.38 & 66.39 & 59.57 \\ \hline
    Cyclist (3D Detection) & 77.15 & 56.49 & 53.37 \\ \hline
    Cyclist (Bird's Eye View) & 81.82 & 60.03 & 56.32 \\ \hline
\end{tabular}
\caption{Performance on KITTI \emph{val} set for pedestrians and cyclists. Model evaluated is Ours (v2).}
\label{tab:val_ped_cyc}
\end{table}

\paragraph{SUN-RGBD} Most previous 3D detection works specialize either on outdoor LiDAR scans where objects are well separated in space and the point cloud is sparse (so that it's feasible for bird's eye projection), or on indoor depth maps that are regular images with dense pixel values such that image CNNs can be easily applied. However, methods designed for bird's eye view may be incapable for indoor rooms where multiple objects often exist together in vertical space. On the other hand, indoor focused methods could find it hard to apply to sparse and large-scale point cloud from LiDAR scans.

In contrast, our frustum-based PointNet is a generic framework for both outdoor and indoor 3D object detection. By applying the same pipeline we used for KITTI data set, we've achieved state-of-the-art performance on SUN-RGBD benchmark (Tab.~\ref{tab:sunrgbd}) with significantly higher mAP as well as much faster (10x-1000x) inference speed.



\begin{table*}[t!]
\small
\centering
\begin{tabular}{l|cccccccccc|c|c}
\hline
          & bathtub & bed & bookshelf & chair & desk & dresser & nightstand & sofa & table & toilet & Runtime  & mAP \\ \hline
DSS~\cite{song2016deep} & 44.2 & 78.8 & 11.9 & 61.2 & 20.5 & 6.4 & 15.4 & 53.5 & 50.3 & 78.9 & 19.55s   & 42.1    \\
COG~\cite{ren2016three} & \textbf{58.3} & 63.7 & 31.8 & 62.2 & \textbf{45.2} & 15.5 & 27.4 & 51.0 & \textbf{51.3} & 70.1 & 10-30min & 47.6 \\
2D-driven~\cite{lahoud20172d} & 43.5 & 64.5 & 31.4 & 48.3 & 27.9 & 25.9 & 41.9 & 50.4 & 37.0 & 80.4 & 4.15s    & 45.1  \\ \hline
Ours (v1) & 43.3 & \textbf{81.1} & \textbf{33.3} & \textbf{64.2} & 24.7 & \textbf{32.0} & \textbf{58.1} & \textbf{61.1} & 51.1 & \textbf{90.9} & 0.12s & \textbf{54.0} \\ \hline
\end{tabular}
\caption{\textbf{3D object detection AP on SUN-RGBD val set.} Evaluation metric is average precision with 3D IoU threshold 0.25 as proposed by~\cite{song2015sun}. Note that both COG~\cite{ren2016three} and 2D-driven~\cite{lahoud20172d} use room layout context to boost performance while ours and DSS~\cite{song2016deep} not. Compared with previous state-of-the-arts our method is 6.4\% to 11.9\% better in mAP as well as one to three orders of magnitude faster.}
\label{tab:sunrgbd}
\end{table*}

\subsection{Architecture Design Analysis} \label{sec:exp_analysis}
In this section we provide analysis and ablation experiments to validate our design choices.

\paragraph{Experiment setup.} Unless otherwise noted, all experiments in this section are based on our v1 model on KITTI data using train/val split as in~\cite{cvpr17chen}. To decouple the influence of 2D detectors, we use ground truth 2D boxes for region proposals and use 3D box estimation accuracy (IoU threshold 0.7) as the evaluation metric. We will only focus on the car category which has the most training examples.

\paragraph{Comparing with alternative approaches for 3D detection.} In this part we evaluate a few CNN-based baseline approaches as well as ablated versions and variants of our pipelines using 2D masks.
In the first row of Tab.~\ref{tab:cnn_vs_pointnet}, we show 3D box estimation results from two CNN-based networks. The baseline methods trained VGG~\cite{simonyan2014very} models on ground truth boxes of RGB-D images and adopt the same box parameter and loss functions as our main method. While the model in the first row directly estimates box location and parameters from vanilla RGB-D image patch, the other one (second row) uses a FCN trained from the COCO dataset for 2D mask estimation (as that in Mask-RCNN~\cite{he2017mask}) and only uses features from the masked region for prediction. The depth values are also translated by subtracting the median depth within the 2D mask. However, both CNN baselines get far worse results compared to our main method.

\begin{figure}[t!]
    \centering
    \includegraphics[width=0.96\linewidth]{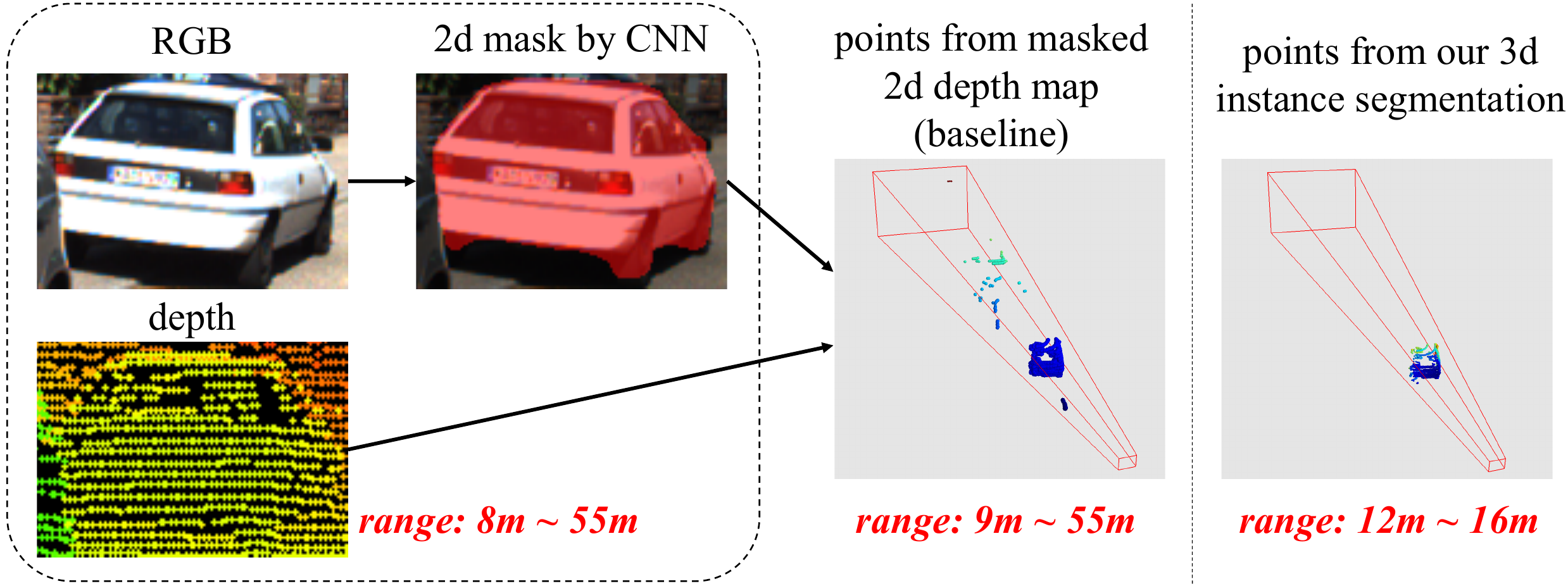}
    \caption{\textbf{Comparisons between 2D and 3D masks.} We show a typical 2D region proposal from KITTI val set with both 2D (on RGB image) and 3D (on frustum point cloud) instance segmentation results. The red numbers denote depth ranges of points.}
    \label{fig:mask2d3d}
\end{figure}

To understand why CNN baselines underperform, we visualize a typical 2D mask prediction in Fig.~\ref{fig:mask2d3d}. While the estimated 2D mask appears in high quality on an RGB image, there are still lots of clutter and foreground points in the 2D mask. 
In comparison, our 3D instance segmentation gets much cleaner result, which greatly eases the next module in finer localization and bounding box regression.


In the third row of Tab.~\ref{tab:cnn_vs_pointnet}, we experiment with an ablated version of frustum PointNet that has no 3D instance segmentation module. Not surprisingly, the model gets much worse results than our main method, which indicates the critical effect of our 3D instance segmentation module. In the fourth row, instead of 3D segmentation we use point clouds from 2D masked depth maps (Fig.~\ref{fig:mask2d3d}) for 3D box estimation. However, since a 2D mask is not able to cleanly segment the 3D object, the performance is more than 12\% worse than that with the 3D segmentation (our main method in the fifth row). On the other hand, a combined usage of 2D and 3D masks -- applying 3D segmentation on point cloud from 2D masked depth map -- also shows slightly worse results than our main method probably due to the accumulated error from inaccurate 2D mask predictions.



\begin{table}[t!]
\small
\centering
\begin{tabular}{c|c|c|c}
\hline
network arch. & mask & depth representation & accuracy \\ \hline
ConvNet        & -    & image & 18.3 \\
ConvNet        & 2D   & image & 27.4 \\ \hline
PointNet       & -    & point cloud & 33.5 \\
PointNet       & 2D   & point cloud & 61.6 \\
PointNet       & 3D   & point cloud & \textbf{74.3} \\
PointNet       & 2D+3D & point cloud & 70.0 \\ \hline
\end{tabular}
\caption{\textbf{Comparing 2D and 3D approaches.} 2D mask is from FCN on RGB image patch. 3D mask is from PointNet on frustum point cloud. 2D+3D mask is 3D mask generated by PointNet on point cloud poped up from 2D masked depth map.}
\label{tab:cnn_vs_pointnet}
\end{table}

\vspace{-0.06in}
\paragraph{Effects of point cloud normalization.} As shown in Fig.~\ref{fig:coordinate}, our frustum PointNet takes a few key coordinate transformations to canonicalize the point cloud for more effective learning. Tab.~\ref{tab:pc_normalization} shows how each normalization step helps for 3D detection. We see that both frustum rotation (such that frustum points have more similar XYZ distributions) and mask centroid subtraction (such that object points have smaller and more canonical XYZ) are critical. In addition, extra alignment of object point cloud to object center by T-Net also contributes significantly to the performance.


\begin{table}[t!]
\small
\centering
\begin{tabular}{c|c|c|c}
\hline
frustum rot. & mask centralize & t-net & accuracy \\ \hline
- & - & - & 12.5 \\
$\surd$ & - & - & 48.1         \\
- & $\surd$ & - & 64.6         \\
$\surd$ & $\surd$ & -& 71.5         \\
$\surd$ & $\surd$ & $\surd$ & \textbf{74.3}         \\ \hline
\end{tabular}
\caption{\textbf{Effects of point cloud normalization.} Metric is 3D box estimation accuracy with IoU=0.7.}
\label{tab:pc_normalization}
\end{table}

\begin{table}[t!]
\small
\centering
\begin{tabular}{c|c|c}
\hline
loss type & regularization & accuracy \\ \hline
regression only & - & 62.9 \\ \hline
cls-reg & - & 71.8    \\ 
cls-reg (normalized) & - & 72.2    \\ \hline
cls-reg (normalized)  & corner loss & \textbf{74.3}  \\ \hline
\end{tabular}
\caption{\textbf{Effects of 3D box loss formulations}. Metric is 3D box estimation accuracy with IoU=0.7.}
\label{tab:corner_loss}

\end{table}

\vspace{-.15in}
\paragraph{Effects of regression loss formulation and corner loss.} In Tab.~\ref{tab:corner_loss} we compare different loss options and show that a combination of ``cls-reg'' loss (the classification and residual regression approach for heading and size regression) and a regularizing corner loss achieves the best result.

The naive baseline using regression loss only (first row) achieves unsatisfactory result because the regression target is large in range (object size from 0.2m to 5m). In comparison, the cls-reg loss and a normalized version (residual normalized by heading bin size or template shape size) of it achieve much better performance. At last row we show that a regularizing corner loss further helps optimization.

\vspace{-.16in}
\subsection{Qualitative Results and Discussion}
\label{sec:exp_viz}
In Fig.~\ref{fig:results} we visualize representative outputs of our frustum PointNet model. We see that for simple cases of non-occluded objects in reasonable distance (so we get enough number of points), our model outputs remarkably accurate 3D instance segmentation mask and 3D bounding boxes. Second, we are surprised to find that our model can even predict correctly posed \emph{amodal} 3D box from partial data (e.g. parallel parked cars) with few points. Even humans find it very difficult to annotate such results with point cloud data only. Third, in some cases that seem very challenging in images with lots of nearby or even overlapping 2D boxes, when converted to 3D space, the localization becomes much easier (e.g. P11 in second row third column).

On the other hand, we do observe several failure patterns, which indicate possible directions for future efforts.
The \emph{first} common mistake is due to inaccurate pose and size estimation in a sparse point cloud (sometimes less than 5 points). We think image features could greatly help esp. since we have access to high resolution image patch even for far-away objects.
The \emph{second} type of challenge is when there are multiple instances from the same category in a frustum (like two persons standing by). Since our current pipeline assumes a single object of interest in each frustum, it may get confused when multiple instances appear and thus outputs mixed segmentation results. This problem could potentially be mitigated if we are able to propose multiple 3D bounding boxes within each frustum.
\emph{Thirdly}, sometimes our 2D detector misses objects due to dark lighting or strong occlusion.  Since our frustum proposals are based on region proposal, no 3D object will be detected given no 2D detection. However, our 3D instance segmentation and amodal 3D box estimation PointNets are not restricted to RGB view proposals. As shown in the supplementary, the same framework can also be extended to 3D regions proposed in bird's eye view.




\vspace{-.11in}
\paragraph{Acknowledgement}
The authors wish to thank the support of Nuro Inc., ONR MURI grant N00014-13-1-0341, NSF grants DMS-1546206 and  IIS-1528025, a Samsung GRO award, and gifts from Adobe, Amazon, and Apple.


\newpage
{\small
\bibliographystyle{ieee}
\bibliography{pcl}
}

\newpage
\appendix
\section{Overview}
This document provides additional technical details, extra analysis experiments, more quantitative results and qualitative test results to the main paper.

In Sec.～\ref{sec:supp_pointnets} we provide more details on network architectures of PointNets and training parameters while Sec.~\ref{sec:supp_rgb_detector} explains more about our 2D detector. Sec.~\ref{sec:supp_bv} shows how our framework can be extended to bird's eye view (BV) proposals and how combining BV and RGB proposals can further improve detection performance. Then Sec.~\ref{sec:supp_more_exp} presents results from more analysis experiments. At last, Sec.~\ref{sec:supp_viz} shows more visualization results for 3D detection on SUN-RGBD dataset.

\section{Details on Frustum PointNets (Sec 4.2, 4.3)}
\label{sec:supp_pointnets}

\subsection{Network Architectures}
\begin{figure*}
    \centering
    \includegraphics{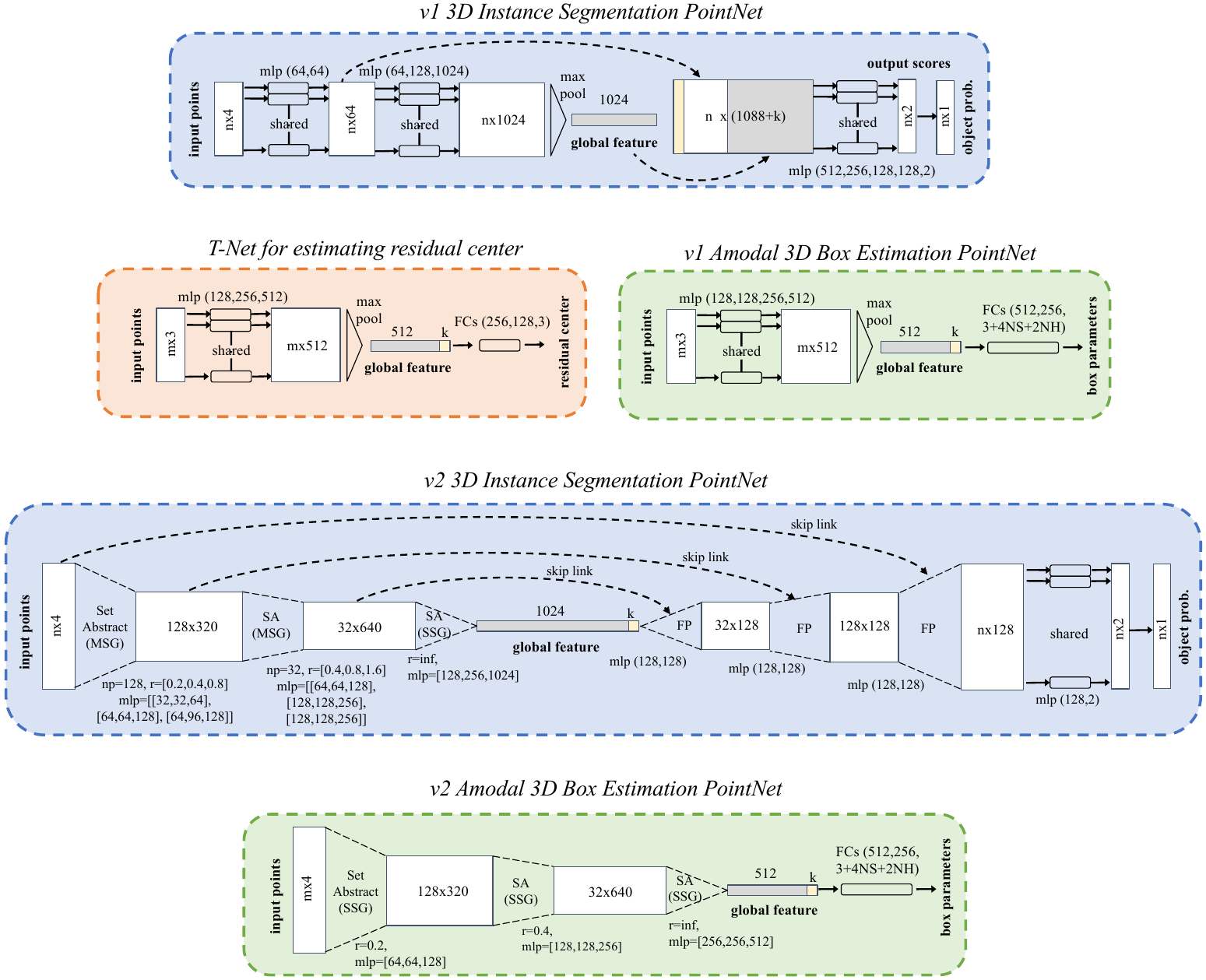}
    \caption{\textbf{Network architectures for Frustum PointNets.} v1 models are based on PointNet~\cite{qi2017pointnet}. v2 models are based on PointNet++~\cite{qi2017pointnetplusplus} set abstraction (SA) and feature propagation (FP) layers. The architecture for residual center estimation T-Net is shared for Ours (v1) and Ours (v2). The colors (blue for segmentaiton nets, red for T-Net and green for box estimation nets) of the network background indicate the coordinate system of the input point cloud. Segmentation nets operate in frustum coordinate, T-Net processes points in mask coordinate while box estimation nets take points in object coordinate. The small yellow square (or bar) concatenated with global features is class one-hot vector that tells the predicted category of the underlying object.}
    \label{fig:arch}
\end{figure*}

We adopt similar network architectures as in the original works of PointNet~\cite{qi2017pointnet} and PointNet++~\cite{qi2017pointnetplusplus} for our v1 and v2 models respectively. What is different is that we add an extra link for class one-hot vector such that instance segmentation and bounding box estimation can leverage semantics predicted from RGB images. The detailed network architectures are shown in Fig.~\ref{fig:arch}.

For v1 model our architecture involves point embedding layers (as shared MLP on each point independently), a max pooling layer and per-point classification multi-layer perceptron (MLP) based on aggregated information from global feature and each point as well as an one-hot class vector. Note that we do not use the transformer networks as in~\cite{qi2017pointnet} because frustum points are viewpoint based (not complete point cloud as in~\cite{qi2017pointnet}) and are already normalized by frustum rotation. In addition to XYZ , we also leverage LiDAR intensity as a fourth channel.

For v2 model we use set abstraction layers for hierarchical feature learning in point clouds. In addition, because LiDAR point cloud gets increasingly sparse as it gets farther, feature learning has to be robust to those density variations. Therefore we used a robust type of set abstraction layers -- multi-scale grouping (MSG) layers as introduced in~\cite{qi2017pointnetplusplus} for the segmentation network. With hierarchical features and learned robustness to varying densities, our v2 model shows superior performance than v1 model in both segmentation and box estimation.

\subsection{Data Augmentation and Training}

\paragraph{Data augmentation} Data augmentation plays an important role in preventing model overfitting. Our augmentation involves two branches: one is 2D box augmentation and the other is frustum point cloud augmentation.

We use ground truth 2D boxes to generate frustum point clouds for Frustum PointNets training and augment the 2D boxes by random translation and scaling. Specifically, we firstly compute the 2D box height ($h$) and width ($w$) and translate the 2D box center by random distances sampled from Uniform[$-0.1w, 0.1w$] and Uniform[$-0.1h, 0.1h$] in u,v directions respectively. The height and width are also augmented by two random scaling factor sampled from Uniform[$0.9, 1.1$].

We augment each frustum point cloud by three ways. First, we randomly sample a subset of points from the frustum point cloud on the fly (1,024 for KITTI and 2,048 for SUN-RGBD). For object points segmented from our predicted 3D mask, we randomly sample 512 points from it (if there are less than 512 points we will randomly resample to make up for the number). Second, we randomly flip the frustum point cloud (after rotating the frustum to the center) along the YZ plane in camera coordinate (Z is forward, Y is pointing down). Thirdly, we perturb the points by shifting the entire frustum point cloud in Z-axis direction such that the depth of points is augmented. Together with all data augmentation, we modify the ground truth labels for 3D mask and headings correspondingly. 

\paragraph{KITTI Training} The object detection benchmark in KITTI provides synchronized RGB images and LiDAR point clouds with ground truth amodal 2D and 3D box annotations for vehicles, pedestrians and cyclists. The training set contains 7,481 frames and an undisclosed test set contains 7,581 frames. In our own experiments (except those for test sets), we follow ~\cite{chen2016monocular,cvpr17chen} to split the official training set to a train set of 3,717 frames and a val set of 3769 frames such that frames in train/val sets belong to different video clips. For models evaluated on the test set we train our model on our own train/val split where around 80\% of the training data is used such that the model can achieve better generalization by seeing more examples.

To get ground truth for 3D instance segmentation we simply consider all points that fall into the ground truth 3D bounding box as object points. Although there are sometimes false labels from ground points or points from other closeby objects (e.g. a person standing by), the auto-labeled segmentation ground truth is in general acceptable.

For both of our v1 and v2 models, we use Adam optimizer with starting learning rate 0.001, with step-wise decay (by half) in every 60k iterations. For all trainable layers except the last classification or regression ones, we use batch normalization with a start decay rate of 0.5 and gradually decay the decay rate to 0.99 (step-wise decay with rate 0.5 in every 20k iterations). We use batch size 32 for v1 models and batch size 24 for v2 models. All three PointNets are trained end-to-end.

Trained on a single GTX 1080 GPU, it takes around one day to train a v1 model (all three nets) for 200 epochs while it takes around three days for a v2 model. We picked the early stopped (200 epochs) snapshot models for evaluation.

\begin{table*}[t!]
\small
\centering
\begin{tabular}{l|ccc||ccc||ccc}
\hline
\multirow{2}{*}{Method} & \multicolumn{3}{c||}{Cars} & \multicolumn{3}{c||}{Pedestrians} & \multicolumn{3}{c}{Cyclists} \\ \cline{2-10} 
                        & Easy  & Moderate  & Hard  & Easy     & Moderate    & Hard    & Easy    & Moderate   & Hard   \\ \hline
SWC & \textbf{90.82} & 90.05 & 80.59 & 87.06 & \textbf{78.65} & 73.92 & \textbf{86.02} & \textbf{77.58} & \textbf{68.44} \\
RRC~\cite{Ren17CVPR} & 90.61 & \textbf{90.22} & \textbf{87.44} & 84.14 & 75.33 & 70.39 & 84.96 & 76.47 & 65.46 \\
Ours & 90.78 & 90.00 & 80.80 & \textbf{87.81} & 77.25 & \textbf{74.46} & 84.90 & 72.25 & 65.14 \\ \hline
\end{tabular}
\caption{\textbf{2D object detection} AP on KITTI \emph{test} set. Evaluation IoU threshold is 0.7. SWC is the first place winner on KITTI leader board for pedestrians and cyclists at the time of submission. Our 2D results are based on a CNN model on monocular RGB images.}
\label{tab:kitti_test_2d}
\end{table*}

\begin{table}[t!]
\centering
\begin{tabular}{l|ccc}
\hline
Subset & Easy    & Moderate    & Hard   \\ \hline
AP (2D) for cars   & 96.48 & 90.31 & 87.63 \\ \hline
\end{tabular}
\caption{\textbf{Our 2D object detection} AP on KITTI \emph{val} set.}
\label{tab:kitti_val_2d_detection}
\end{table}

\paragraph{SUN-RGBD Training} The data set consists of 10,355 RGB-D images captured from various depth sensors for indoor scenes (bedrooms, dining rooms etc.). We follow the same train/val splits as \cite{song2015sun,ren2016three} for experiments. The data augmentation and optimization parameters are the same as that in KITTI.

As to auto-labeling of instance segmentation mask, however, data quality is much lower than that in KITTI because of strong occlusions and tight arrangement of objects in indoor scenes (see Fig.~\ref{fig:sunrgbd_viz} for some examples). Nonetheless we still consider all points within the ground truth boxes as object points for our training. For 3D segmentation we get only a 82.7\% accuracy compared to around 90\% in KITTI. Due to the heavy noise in segmentation mask label, we choose to only train and evaluate on v1 models that has more strength in global feature learning than v2 ones. For future works, we think higher quality in 3D mask labels can greatly help the instance segmentation network training.

\section{Details on RGB Detector (Sec 4.1)}
\label{sec:supp_rgb_detector}

For 2D RGB image detector, we use the encoder-decoder structure (e.g. DSSD~\cite{fu2017dssd}, FPN~\cite{lin2016feature}) to generate region proposals from multiple feature maps using focal loss~\cite{lin2017focal} and use Fast R-CNN~\cite{girshick2015fast} to predict final 2D detection bounding boxes from the region proposals.

To make the detector faster, we take the reduced VGG~\cite{simonyan2014very} base network architecture from SSD~\cite{liu2016ssd}, sample half of the channels per layer and change all max pooling layers to convolution layers with $3\times3$ kernel size and stride of 2. Then we fine-tune it on ImageNet CLS-LOC dataset for 400k iterations with batch size of 260 on 10 GPUs. The resulting base network architecture has about 66.7\% top-1 classification accuracy on the CLS-LOC validation dataset and only needs about 1.2ms to process a $224\times224$ image on a NVIDIA GTX 1080.

We then add the feature pyramid layers~\cite{lin2016feature} from conv3\_3, conv4\_3, conv5\_3, and fc7, which are used to predict region proposals with scales of 16, 32, 64, 128 respectively. We also add an extra convolutional layer (conv8) which halves the fc7 feature map size, and use it to predict proposals with scale of 256. We use 5 different aspect ratios \{$\frac{1}{3}$, $\frac{1}{2}$, 1, 2, 3\} for all layers except that we ignore \{$\frac{1}{3}$, 3\} for conv3\_3. Following SSD, we also use normalization layer on conv3\_3, conv4\_3, and conv5\_3 and initialize the norm 40. For Fast R-CNN part, we extract features from conv3\_3, conv5\_3, and conv8 for each region proposal and concatenate all the features to predict class scores and further adjust the proposals. We train this detector from COCO dataset with $384\times384$ input image and have achieved 35.5 mAP on the COCO minival dataset, with only 10ms processing time for a $384\times384$ image on a single GPU.

Finally, we fine-tune the detector on car, people, and bicycle from COCO dataset, and have achieved 48.5, 44.1, and 40.1 for these three classes on COCO. We take this model and further fine-tune it on car, pedestrian, and cyclist from KITTI dataset. The final model takes about 30ms to process a $384\times1280$ image. To increase the recall of the detector, we also do detection from the center crop of the image besides the full image, and then merge the detections using non-maximum suppression.

Tab.~\ref{tab:kitti_test_2d} shows our detector's AP (2D) on KITTI test set. Our detector has achieved competitive or better results than current leading players on KITTI leader board. We've also reported our AP (2D) on val set in Tab.~\ref{tab:kitti_val_2d_detection} for reference.


\section{Bird's Eye View PointNets (Sec 5.3)}
\label{sec:supp_bv}

In this section, we show that our 3D detection framework can also be extended to using bird's eye view proposals, which adds another orthogonal proposal source to achieve better overall 3D detection performance. We evaluate the results of car detection using LiDAR bird's eye view only proposals + point net (Ours(BV)), and combine frustum point net and bird's eye view point net using 3D non-maximum suppression (NMS) (Ours(Frustum + BV)). The results are shown in Table~\ref{tab:td_re}.

\paragraph{Bird's Eye View Proposal} Similar to MV3D~\cite{cvpr17chen} we use point features such as height, intensity and density, and train the bird's eye view 2D proposal net using the standard Faster-RCNN~\cite{ren2015faster} structure. The net outputs axis-aligned 2D bounding boxes in the bird's eye view. In detail, we discretize the projected point clouds into 2D grids with resolution of $0.1$ meter and with the depth and width range $0 ~ 60$ meters, which gives us the $600\times600$ input size. For each cell, we take the intensity and the density of the highest point and divide the heights into $7$ bins with the height of the highest point in each bin, which gives us $9$ channels in total. In Faster R-CNN, we use the VGG-16~\cite{simonyan2014very} with $3$ anchor scales ($16, 32, 48$) and $3$ aspect ratios ($\frac{1}{2}, 1, 2$). We train RPN and Fast R-CNN together using the approximate joint training.

To combine 3D detection boxes from frustum PointNets and the bird's eye view PointNets, we use 3D NMS with IoU threshold $0.8$. We also apply a weight (0.5) to 3D boxes from BV PointNets since it is a weaker detector compared with our frustum one.

\paragraph{Bird's Eye View (BV) PointNets} Similar to Frustum PointNets that take point cloud in frustum, segment point cloud and estimate amodal bounding box, we can apply PointNets to points in bird's eye view regions. Since bird's eye view is based on orthogonal projection, the 3D space specified by a BV 2D box is a 3D cuboid (cut by minimum and maximum height) instead of a frustum.

\paragraph{Results} Tab.~\ref{tab:td_re} (Ours BV) shows the APs we get by using bird's eye view proposals only (without and RGB information). We compare with two previous LiDAR only methods (VeloFCN~\cite{li2016vehicle} and MV3D (BV+FV)~\cite{cvpr17chen}) and show that our BV proposal based detector greatly outperforms VeloFCN on all cases and outperforms MV3D (BV+FV) on moderate and hard cases by a significant margin.

\begin{figure}[h!]
    \centering
    \includegraphics[width=\linewidth]{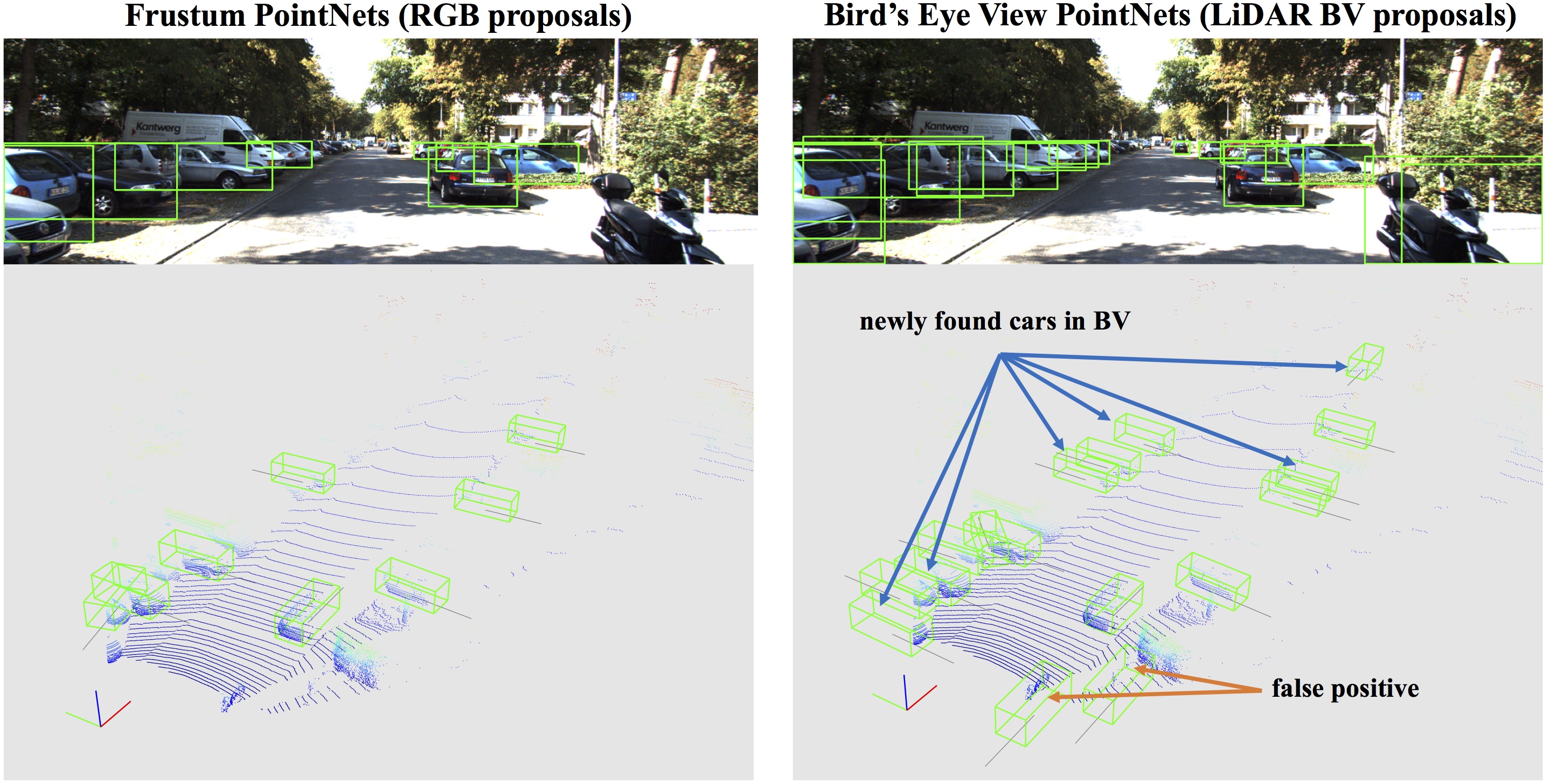}
    \caption{\textbf{Comparing Frustum PointNets and BV PointNets.} This is a scene with lots of parallel parking cars (sample 5595 from val set). \emph{Left} column shows 2D boxes from our 2D detector in image and 3D boxes from our Frustum PointNets in point cloud. \emph{Right} column shows 3D boxes from BV PointNets in point cloud and the 2D boxes (projected from the 3D detection boxes) in image. Note that 2D detection boxes from Ours (Frustum) that have box height less than 25 pixels or contain no LiDAR points in the frustum are not shown in the image.}
    \label{fig:frustum_vs_bv}
\end{figure}

More importantly, we show in the last row of Tab.~\ref{tab:td_re} that bird's eye view and RGB view proposals can be combined to achieve an even better performance (\emph{3.8\%} AP improvement on hard cases).
Fig.~\ref{fig:frustum_vs_bv} gives an intuitive explanation of why bird's eye view proposals could help. In the sample frame shown: while our 2D detector misses some highly occluded cars (Fig.~\ref{fig:frustum_vs_bv}: left RGB image), bird's eye view based RPN successfully detects them (Fig.~\ref{fig:frustum_vs_bv}: blue arrows in right LiDAR image).

\begin{table}[h!]
    \centering
    \begin{tabular}{l|ccc}
    \hline
        Method & Easy & Moderate & Hard \\ \hline
        VeloFCN~\cite{li2016vehicle} & 15.20 & 13.66 & 15.98 \\
        MV3D~\cite{cvpr17chen} (BV+FV) & 71.19 & 56.60 & 55.30 \\ \hline
        Ours (BV) &  69.50 & 62.30 & 59.73 \\
        Ours (Frustum) & \textbf{83.76} & \textbf{70.92} & 63.65 \\
        Ours (Frustum + BV) & \textbf{83.76} & 70.91 & \textbf{67.47} \\ \hline
    \end{tabular}
    \caption{\textbf{3D object detection} AP on KITTI \emph{val} set. By using both proposals from RGB view (frustum) and bird's eye view (BV), we see a significant improvement in 3D AP (\emph{3.82\%}) on hard cases compared with our frustum only method. Ours (Frustum) here is the Ours (v2) in the main paper using PointNet++ architectures.}
    \label{tab:td_re}
\end{table}

\section{More Experiments (Sec 5.2)}
\label{sec:supp_more_exp}

\subsection{Effects of PointNet Architectures}
Table~\ref{tab:v1v2} compares PointNet~\cite{qi2017pointnet} (v1) and PointNet++~\cite{qi2017pointnetplusplus} (v2) architectures for instance segmentation and amodal box estimation. The v2 model outperforms v1 model on both tasks because 1) v2 model learns hierarchical features that are richer and more generalizable; 2) v2 model uses multi-scale feature learning that adapts to varying point densities. Note that the ours (v1) model corresponds to first row of Table~\ref{tab:v1v2} while the ours (v2) links to the last row.

\begin{table}[h!]
\centering
\begin{tabular}{cc|cc}
\hline
seg net & box net & seg acc. & box acc. \\ \hline
v1   & v1   & 90.6 & 74.3  \\
v2   & v1   & \textbf{91.0} & 74.7 \\
v1   & v2   & 90.6 & 76.0 \\
v2   & v2   & \textbf{91.0}  & \textbf{77.1} \\ \hline
\end{tabular}
\caption{\textbf{Effects of PointNet architectures.} Metric is 3D box estimation accuracy with IoU=0.7.}
\label{tab:v1v2}
\end{table}

\subsection{Effects of Training Data Size}

\begin{figure}[h!]
    \centering
    \includegraphics[width=0.6\linewidth]{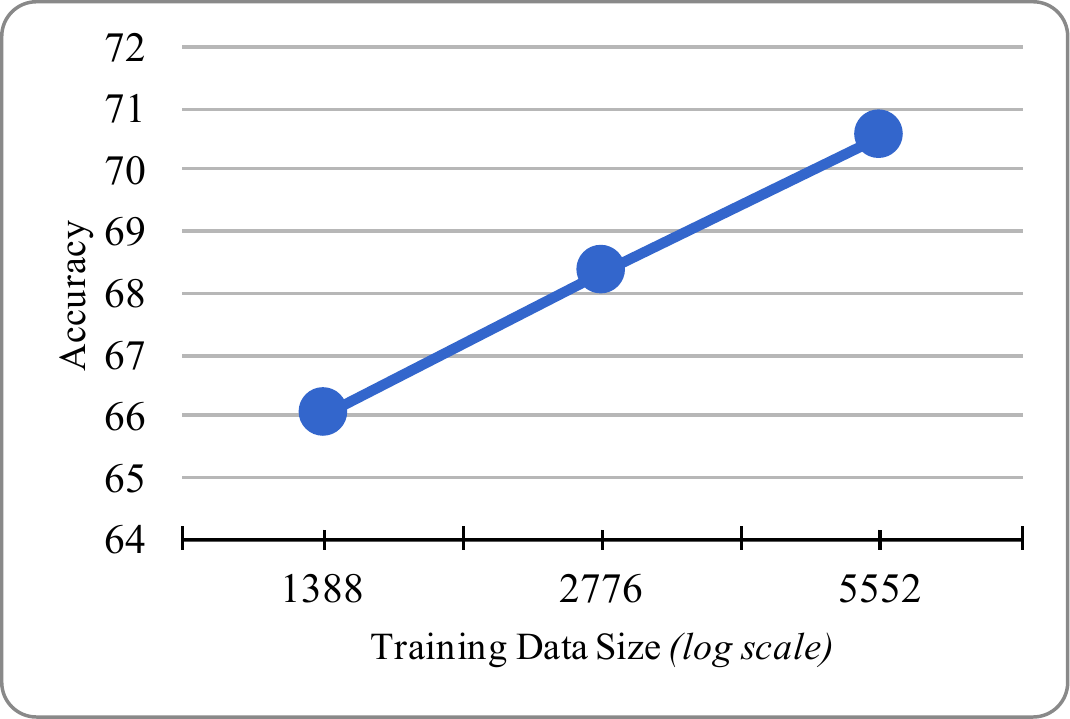}
    \caption{\textbf{Effects of training data size.} Evaluation metric is 3D box estimation accuracy (IoU threshold 0.7). We see a clear trend of linear improvement in accuracy with exponential growth of training data size.}
    \label{fig:training_data_size}
\end{figure}

Recently \cite{sun2017revisiting} observed linear improvement in performance of deep learning models with exponential growth of data set size. In our Frustum PointNets we observe similar trend (Fig.~\ref{fig:training_data_size}). This trend indicates a promising performance potential of our methods with larger datasets.

We train three separate group of Frustum PointNets on three sets of training data and then evaluate the model on a fixed validation set (1929 samples). The three data points in Fig.~\ref{fig:training_data_size} represent training set sizes of 1388, 2776, 5552 samples (0.185x, 0.371x, 0.742x of the entire trainval set) respectively. We augment the training data such that the total amount of samples are the same for each of the three cases (20x, 10x and 5x augmentation respectively). The training set and validation set are chosen such that they don't share frames from the same video clips.

\subsection{Runtime and Model Size}
\label{sec:runtime}
In Table~\ref{tab:runtime}, we show decomposed runtime cost (inference time) for our frustum PointNets (v1 and v2). The evaluation is based on TensorFlow~\cite{abadi2016tensorflow} with a NVIDIA GTX 1080 and a single CPU core. While for v1 model frustum proposal (with CNN and backprojection) takes the majority time, for v2 model since a PointNet++~\cite{qi2017pointnetplusplus} model with multi-scale grouping is used, computation bottleneck shifts to instance segmentation. Note that we merge batch normalization and FC/convolution layers for faster inference (since they are both linear operation with multiply and sum), which results in close to 50\% speedup for inference.

CNN model has size 28 MB. v1 PointNets have size 19MB. v2 PointNets have size 22MB. The total size is therefore 47MB for v1 model and 50MB for v2 model.

\begin{table}[h!]
\small
\centering
\begin{tabular}{c|ccc|c}
\hline
Model & Frustum Proposal & 3D Seg &  Box Est. & Total \\ \hline
v1 & 60 ms & 18 ms & 10 ms & 88 ms \\ 
v2 & 60 ms & 88 ms & 19 ms & 167 ms \\ \hline
\end{tabular}
\caption{\textbf{3D detector runtime.} Thirty-two region proposals used for frustum-based PointNets. 1,024 points are used for instance segmentation and 512 points are used for box estimation.}
\label{tab:runtime}
\end{table}

\section{Visualizations for SUN-RGBD (Sec 5.1)}
\label{sec:supp_viz}

\begin{figure*}[t!]
    \centering
    \includegraphics[width=0.95\linewidth]{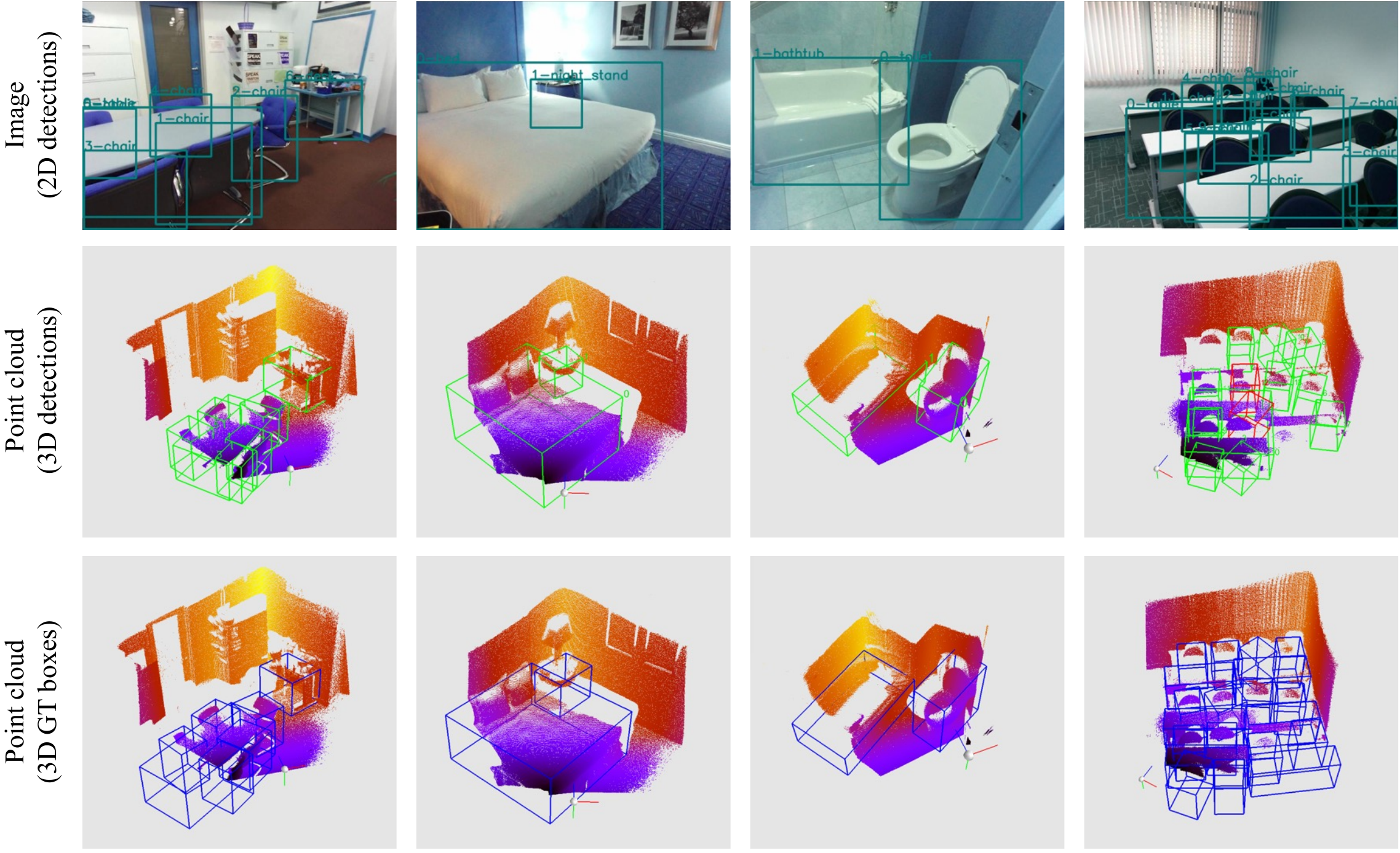}
    \caption{\textbf{Visualization of Frustum PointNets results on SUN-RGBD val set.} \emph{First row:} RGB image with 2D detection boxes. \emph{Second row:} point cloud popped up from depth map and predicted amodal 3D bounding boxes (the numbers beside boxes correspond to 2D boxes on images). Green boxes are true positive. Red boxes are false positives. False negatives are not visualized. \emph{Third row:} point cloud popped up from depth map and ground truth amodal 3D bounding boxes.}
    \label{fig:sunrgbd_viz}
\end{figure*}

In Fig.~\ref{fig:sunrgbd_viz} we visualize some representative detection results on SUN-RGBD data. We can see that compared with KITTI LiDAR data, depth images can be popped up to much more dense point clouds. However even with such dense point cloud, strong occlusions of indoor objects as well as the tight arrangement present new challenges for detection in indoor scenes.

\begin{figure*}[t!]
    \centering
    \includegraphics[width=\linewidth]{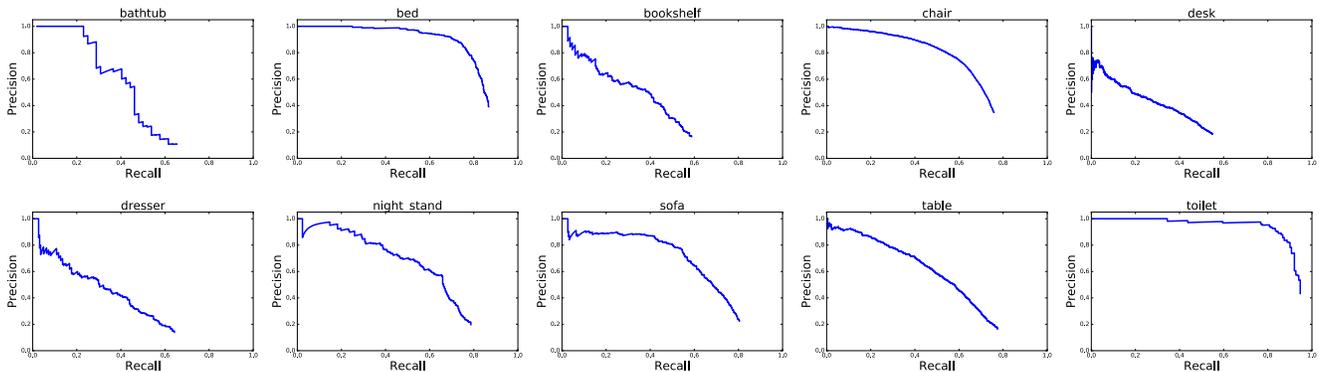}
    \caption{\textbf{Precision recall (PR) curves} for 3D object detection on SUN-RGBD val set.}
    \label{fig:sunrgbd_ap}
\end{figure*}

In Fig.~\ref{fig:sunrgbd_ap} we report the 3D AP curves of our Frustum PointNets on SUN-RGBD val set. 2D detection APs of our RGB detector are also provided in Tab.~\ref{tab:kitti_val_2d_detection} for reference. 


\begin{table*}[]
    \centering
    \begin{tabular}{l|cccccccccc|c}
    \hline
    Category & bathtub & bed & bookshelf & chair & desk & dresser & nightstand & sofa & table & toilet & mean \\ \hline
         AP (2D) & 81.3 & 56.7 & 67.2 & 64.1 & 77.8 & 33.3 & 37.2 & 57.4 & 49.9 & 43.5 & 50.3 \\ \hline
         AP (3D) & 43.3 & 81.1 & 33.3 & 64.2 & 24.7 & 32.0 & 58.1 & 61.1 & 51.1 & 90.9 & 54.0 \\ \hline
    \end{tabular}
    \caption{\textbf{2D and 3D object detection} AP on SUN-RGBD val set. 2D IoU threshold is 0.5. Note that on some categories we get higher 3D AP (displayed in the table as well, the same results as in main paper) than 2D AP because our network is able to recover 3D geometry from very partial scan and is also due to a more loose 3D IoU threshold (0.25) in SUN-RGBD 3D AP evaluation.}
    \label{tab:my_label}
\end{table*}

\end{document}